\RequirePackage{fix-cm}
\documentclass[twocolumn]{svjour3}          
\smartqed  
\usepackage{graphicx}
\usepackage{bbding}
\usepackage{float}
\usepackage{multirow}
\usepackage{color}
\usepackage{lineno}
\usepackage{bm}
\usepackage{amsfonts}
\begin{document}

\title{Data-Driven Prediction of Dynamic Interactions Between Robot Appendage and Granular Material }


\author{Guanjin Wang \textsuperscript{+} \and 
       Xiangxue Zhao \textsuperscript{+} \and Shapour Azarm \and Balakumar Balachandran
}


\institute{\textsuperscript{+} These authors contributed equally.\\
            Guanjin Wang 
            \at
              The Institute for Data Intensive Engineering and Science\\
              Hopkins Extreme Materials Institute\\
              Department of Mechanical Engineering\\
              Johns Hopkins University, Baltimore, MD 21218 \\
              \email{gjwang@terpmail.umd.edu}             
           \and
            Xiangxue Zhao 
            \at
              Department of Mechanical Engineering\\
              University of Maryland, College Park, MD 20742 \\
              \email{xiangxuezhao@gmail.com}             
           \and
            Shapour Azarm and B.Balachandran \at
              Department of Mechanical Engineering\\
              University of Maryland, College Park, MD 20742 \\
              \email{azarm,balab@umd.edu}           
           \and
}

\date{Received: date / Accepted: date}

\maketitle

\begin{abstract}
Legged and hybrid locomotion have advantages over wheeled motion on compliant surfaces. With physics-based, high-fidelity simulations, complex physics associated with robot navigation on challenging terrain can be predicted. However, prediction can be improved if the data obtained from physics-based simulations are supplemented with the data associated with physical experimental measurements. In this paper, an alternative data-driven modeling approach has been proposed and employed to gain fundamental insights into robot motion interaction with granular terrain at certain length scales. The approach is based on an integration of dimension reduction (Sequentially Truncated Higher-Order Singular Value Decomposition), surrogate modeling (Gaussian Process), and data assimilation techniques (Reduced Order Particle Filter). This approach can be used online and is based on offline data, obtained from the offline collection of high-fidelity simulation data and a set of sparse experimental data. The results have shown that orders of magnitude reduction in computational time can be obtained from the proposed data-driven modeling approach compared with physics-based high-fidelity simulations. With only simulation data as input, the data-driven prediction technique can generate predictions that have comparable accuracy as simulations. With both simulation data and sparse physical experimental measurement as input, the data-driven approach with its embedded data assimilation techniques has the potential in outperforming only high-fidelity simulations for the long-horizon predictions. In addition, it is demonstrated that the data-driven modeling approach can also reproduce the scaling relationship recovered by physics-based simulations for maximum resistive forces, which may indicate its general predictability beyond a case-by-case basis. The results are expected to help robot navigation and exploration in unknown and complex terrains during both online and offline phases. 

\end{abstract}
\section{Introduction}
\label{intro}
The successful landing of Mars exploration rovers has enabled the return of millions of Mars surface images for important scientific discoveries and observations about the Martian environment. The investigations of the mobility of these space robots and spacecraft on sandy terrain are important to maintaining their operating service in the natural environment \cite{Calvin}. Legged locomotion has superior mobility in soft, hostile, and challenging terrains over wheeled locomotion, which suffers from lack of traction and can result in slipping, sinkage, or even permanent immobility \cite{Li_1}. In order to improve motion effectiveness, the behavior of legged locomotion needs to be addressed in design and operation for robotic systems, and thus, effective prediction is needed to estimate legged dynamic behavior under different robotic appendage designs and operational conditions.  

Physics-based modeling can serve as a fairly accurate approach to simulate interactions between robotic appendages and granular terrain at laboratory and field scales (\cite{Wang_1}-\cite{Wang_2}, \cite{Zhang_2}). In \cite{Wang_1}, a high-fidelity simulation framework was developed to capture legged locomotion by modeling small light vehicle terrain interaction, sharp edge contact, or irregular surfaces, which has advantages over the classical Bekker terramechanics theory which is limited to large heavy vehicle terrain interaction. However, high-fidelity simulations are usually computationally expensive and do not always account for uncertainty as in a real environment. Therefore, physics-based simulations cannot always satisfy the need for online operation, motion planning, and control, which requires fast prediction and information fusion in a complex environment \cite{Wang_2}. On the other hand, data-driven modeling approaches can be used to provide a just-in-time prediction of legged locomotion behavior for online robot navigation in natural terrain. However, data-driven models which are based on only one source of data (obtained from either simulation or experiment data) may not be sufficient, as simulations might not be fast for online operation, and physical experiment measurements can be noisy and very limited due to the high-expense in data collection process (\cite{Zhao_1}, \cite{Zhao_2}).

Dynamic Data-Driven Application Systems (DDDAS) can be a promising candidate to solve the issue. DDDAS enables incorporating additional data into an executable application (e.g., simulation), where data can be archival offline or collected online \cite{Darema}. DDDAS has many applications, such as forecasting weather \cite{Fisher}, prediction of fire spread and propagation \cite{Mandel}, identification of airborne contaminants \cite{Akçelik}, operation and control of unmanned vehicles \cite{Peng}, and image processing \cite{Uzkent}. However, the potential of DDDAS has not been explored in terramechanics. 

To advance the field of terramechanics, the DDDAS concept is used in the current investigation.The objective of this article is to develop a data-driven approach to predict the interaction of the robotic appendage between the granular material using data from high-fidelity simulations and physical experiments. To address the unique challenges of a limited number of training data, the computational cost of online data processing, and the uncertainty of data, the previous work of the authors \cite{Zhao_1} is further extended in this work to study robotic interaction by varying the design of the appendix, the operation conditions, and the temporal angles. For data acquisition, high-fidelity simulations are used to generate data for system response to feed and train the machine learning algorithms.  Sequentially Truncated Higher-Order Singular Value Decomposition (ST-HOSVD) (\cite{De Lathauwer},\cite{Vannieuwenhoven}) is used for decomposition and dimensional reduction due to its lower computational cost and ability to address high-dimensional data. Then Gaussian Process Regression (GPR) (\cite{Deisenroth}) is employed to construct a surrogate model and forecast the evolution of individual components of the basis. Finally, a reduced-order Particle Filter (\cite{Moradkhani_1},\cite{Moradkhani_2}) is employed to dynamically improve model prediction with the incorporation of experimental measurement at a lower computational cost. 

The rest of this article is organized as follows. In Section 2, the statement of the problem and the associated assumptions are provided. Then, the proposed approach is detailed in Section 3. Next, in Section 4, to demonstrate the effectiveness of the developed approach, a terra-mechanics case study involving temporal-like prediction for robot appendage interaction with granular terrain is presented. Then, it is shown that the scaling relationship recovered by physics-based simulations for maximum resistive forces can also be predicted by the data-driven modeling approach. Finally, discussions and concluding remarks are provided in Section 5. 
\section{Problem Statement and Assumptions}
\label{Problem Statement and Assumptions}

\begin{figure}[b]
  \includegraphics[width=0.5\textwidth]{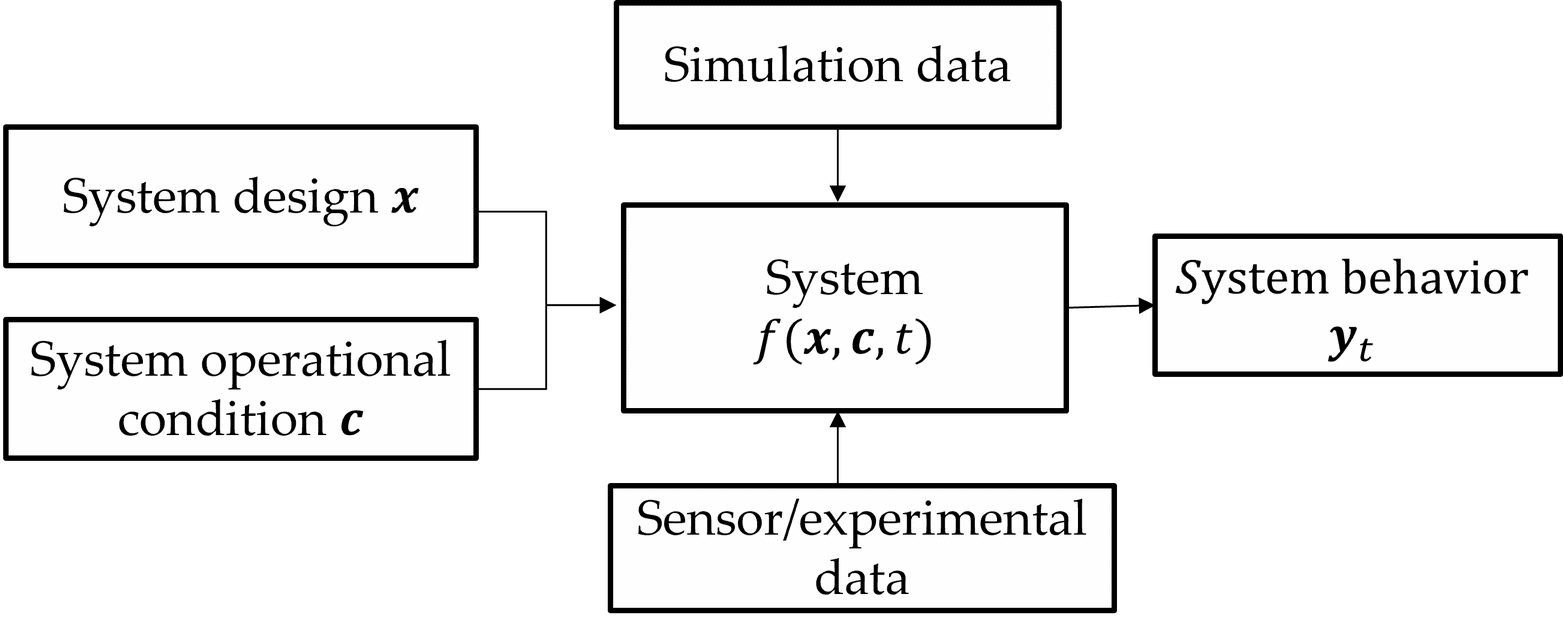}
\caption{A block diagram of legged locomotion prediction problem.}
\label{Fig1}       
\end{figure}

In Fig.\ref{Fig1}, a block diagram of the legged locomotion prediction problem is depicted. The objective is to predict drag and lift forces $\mathbf{y}_\theta$ along temporal-like quantity $\theta$ when the robotic appendage interacts with granular material, which is given by the function of $\mathbf{y}_\theta =f(\mathbf{x},c,\theta)$. Here, $\theta$ is the angle between the robot leg and the vertical direction; $\mathbf{x} \in \mathbb{R}^m$ is a vector of variables that define the design of robotic devices, which can be the length, width, depth and curvature of robotic legs. Examples of robotic appendage design can be reversed C-leg, reversed L-leg, flat leg, L-leg, and C-leg because they represent different leg morphologies in real life. On the other hand, $c \in \mathbb{R}^n$ refers to an operation condition, which is the clockwise stride frequency when robotic appendages move through the granular material. 

\begin{figure*}[htbp]
	\includegraphics[width=1\textwidth]{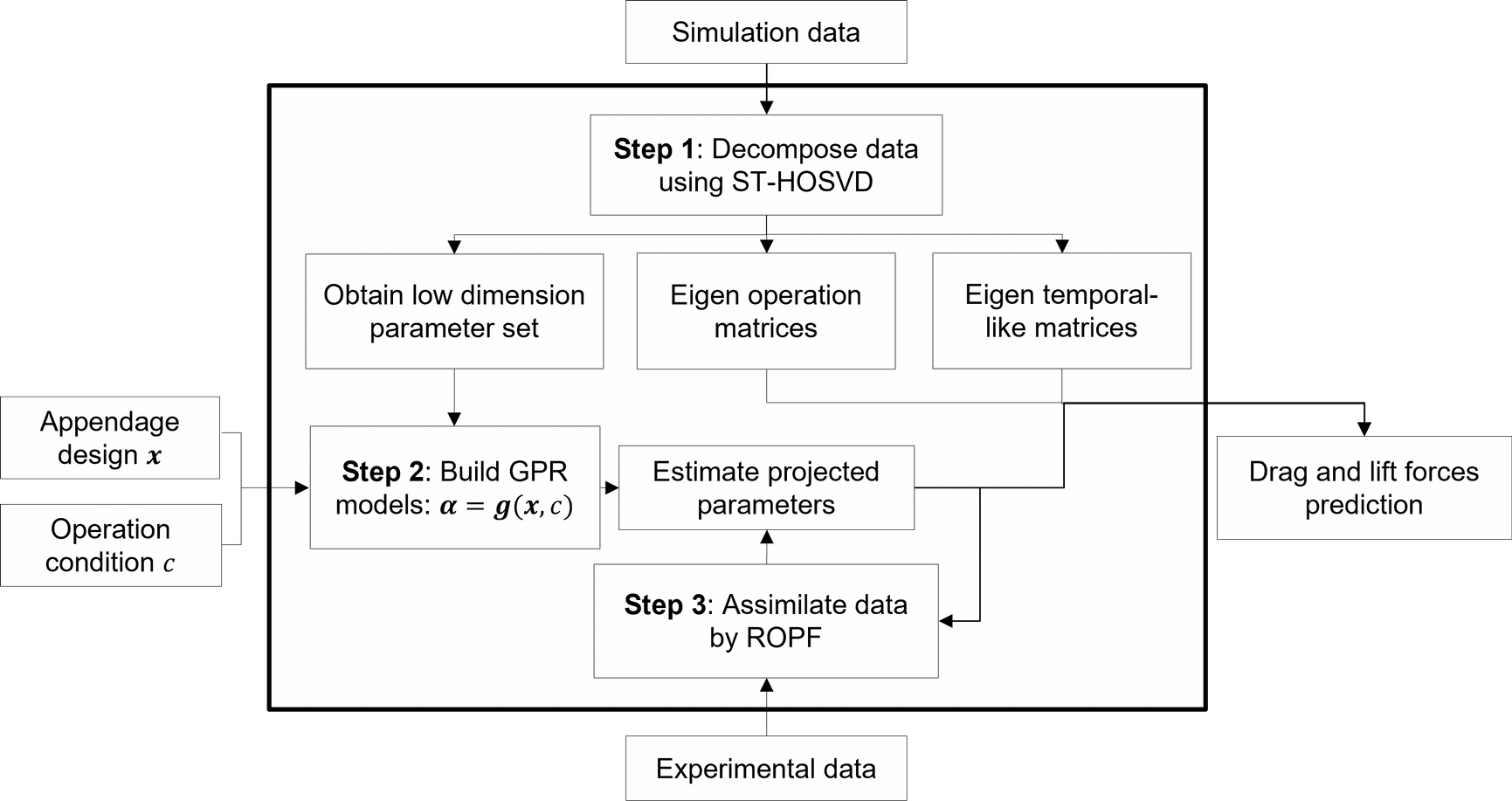}
	\caption{\label{Fig2}The data-driven modeling approach.} 
\end{figure*}

In this problem, two sources of data are available from simulation and physical experiments. First, physics-based simulations \cite{Wang_1} can be used to collect a set of data that are conducted on different combinations of robotic appendage designs and operation conditions. Among the simulated data set on different appendage design and operations, only a very limited subset of scenarios has experiment data. In addition, the experiment data are discrete and subjected to noisy measurement. The goal is to provide an effective prediction for different robot appendage design and operations through the integration of simulation and experiment data.

\section{Simulation Framework and Data Generation }

The experimental data used in the data assimilation are from literature \cite{Li_2}. For the chosen robot -terrain interaction, the physics-based simulations from the authors’ earlier work \cite{Wang_1} are employed to generate the training and testing data for the data driven model. The basic governing equations for granular material are a combination of the conservation laws in fluid mechanics and the constitutive law in solid mechanics. Following the standard notation for continuum mechanics, the momentum equation is given by
\begin{equation}
\begin{array}{lcl}
\displaystyle \frac{D\mathbf{v}}{Dt} &=& \displaystyle\frac{1}{\rho} (\nabla\cdot \boldsymbol{\sigma}) + \mathbf{b}\\[6pt]
\end{array}\vspace*{-12pt}
\end{equation}
\vspace*{-3pt}
the conservation of mass is described by,
\begin{equation}
\begin{array}{lcl}
\displaystyle \frac{D\rho}{Dt} &=& \displaystyle - \rho tr\mathbf{L}\\[6pt]
\end{array}\vspace*{-12pt}
\end{equation}
\vspace*{-3pt}
where $\mathbf{v}$ is the velocity vector, $\rho$ is the material density, $\mathbf{b}$ is the specific body force, and $\boldsymbol{\sigma}$ is the Cauchy stress tensor. $\mathbf{L}$ is the spatial velocity gradient of the form $\displaystyle \mathbf{L} = \displaystyle \nabla\mathbf{v}$ and its trace is $tr\mathbf{L}$. The strain rate tensor is defined as $\displaystyle \mathbf{D} = \displaystyle\frac{1}{2}(\mathbf{L}+\mathbf{L}^{T})$. 
The total material time derivative is defined as 
\begin{equation}
\begin{array}{lcl}
\displaystyle \frac{D()}{Dt} &=& \displaystyle \frac{\partial()}{\partial t}+\mathbf{v}\cdot\frac{\partial()}{\partial \mathbf{x}}\\[6pt]
\end{array}\vspace*{-12pt}
\end{equation}
\vspace*{-3pt}

A constitutive viscoplastic relationship developed for the dense material is adopted from the earlier work of the authors \cite{Wang_1}, where all other detailed descriptions of the parameters can be found.

\begin{equation}\label{constitutive}
\begin{array}{lcl}
\displaystyle  \boldsymbol{\sigma} &=& \displaystyle  \left\{ \begin{array}{lcl} {\displaystyle - P\mathbf{I}+\frac{(\mu(I) P+c)\mathbf{D}}{\vert\mathbf{D}\vert} } & \mbox{for}&  I> 0.001 \\[12pt] \displaystyle - P\mathbf{I}+\frac{(\mu(I) P+c)\mathbf{E}}{\vert\mathbf{E}\vert} & \mbox{for} &I \leq 0.001 \end{array}\right.
\end{array}\vspace*{-12pt}
\end{equation}
\vspace*{-3pt}
Here, in Eq. (\ref{constitutive}), $P$ is the hydro-static pressure determined by using the equation of state, and $\mathbf{I}$ is the identity tensor. The second term in both equations of Eq. (\ref{constitutive}) is the deviatoric shear stress, where c is the representative of cohesion which is 0 for dry granular materials, $I$ is the inertial number. $\vert\mathbf{D}\vert = \sqrt{\frac{1}{2}\mathbf{D:D}}$ is the second invariant of the strain rate tensor $\mathbf{D}$. $\vert\mathbf{E}\vert = \sqrt{\frac{1}{2}\mathbf{E:E}}$ is the second invariant of the strain tensor $\mathbf{E}$, where $\mathbf{E}= \frac{1}{2}(\nabla\mathbf{x}+ \nabla\mathbf{x}^T)$.  $\mu$ is the friction coefficient from reference \cite{Jop} and reads as
\begin{equation}
\begin{array}{lcl}
\displaystyle \mu(I)  &=& \displaystyle \mu_1+\frac{\mu_2-\mu_1}{I_o/I+1}\\[6pt]
\end{array}\vspace*{-12pt}
\end{equation}
\vspace*{-3pt}
\noindent Here,  $\mu_1$ is the friction coefficient in the quasi-static limit, $\mu_2$ is a limiting coefficient for granular material in the dynamic flow regime.

In these experimentally validated simulations, the conservation laws and rheology of granular material are implemented in a Smoothed Particle Hydrodynamics (SPH) framework \cite{Monaghan}, which can accurately capture the large deformation of the granular terrain response during robot leg interaction without grid distortion. The SPH method, which is a mesh free method, is employed for the discretization of the solution domain. The particle approximation for any field quantity f(x) at point i is expressed as,
\begin{equation}
\begin{array}{lcl}
\displaystyle f(\mathbf{x_i}) = \displaystyle \sum\limits_{j=1}^{N}{\frac{m_j}{\rho_j}f(\mathbf{x_j})W_{ij}}  \\[6pt]
\end{array}\vspace*{-12pt}
\end{equation}
\vspace*{-3pt}
The corresponding derivative is given by,
\begin{equation}
\begin{array}{lcl}
\displaystyle  \nabla f(\mathbf{x_i}) = \displaystyle \sum\limits_{j=1}^{N}{\frac{m_j}{\rho_j}f(\mathbf{x_j})\mathbf{\nabla_iW_{ij}}}\\[6pt]
\end{array}\vspace*{-12pt}
\end{equation}
\vspace*{-3pt}

For density initialization , the density calculated by using the summation density before the simulation is executed. For density reinitialization, the Shepard filter is applied every 10 time steps. For density evolution at each time step, the continuity density is calculated from the continuity equation and the  $\delta$ SPH scheme \cite{Marrone} to reduce high-frequency oscillations, 
\begin{equation}
\begin{array}{lcl}
\displaystyle \frac{D\rho_i}{Dt}&=&\displaystyle \sum_{j}{m_j(\mathbf{v_i}-\mathbf{v_j})\cdot\nabla_i\mathbf{W_{ij}}}+\displaystyle \sum_{j}{\psi_{ij}\mathbf{r_{ij}}\nabla_i\mathbf{W_{ij}}}\\[6pt]
\end{array}\vspace*{-12pt}
\end{equation}
\vspace*{-3pt}

The momentum equation is approximated as  
\begin{equation}\label{SPHmomentum}
\begin{array}{lcl}
\displaystyle \frac{D\mathbf{v_i}}{Dt}&=&\displaystyle \sum_{j}{m_j\left(\frac{\boldsymbol{\sigma_i}}{\rho_{i}^{2}}+\frac{\boldsymbol{\sigma_j}}{\rho_{j}^{2}}-\Pi_{ij}\right)\cdot\nabla_i\mathbf{W_{ij}}}+\mathbf{b}\\[6pt]
\end{array}\vspace*{-12pt}
\end{equation}
\vspace*{-3pt}
where  $\Pi$ is the artificial viscosity adopted to improve numerical stability and avoid inter-particle penetration. The Verlet–Leapfrog algorithm is then used to integrate the discretized governing equations to obtain the solution field.

\section{Approach}
\label{sec:2}
The approach is an extension of the authors’ previous work \cite{Zhao_1}. In this section, the adaptation of this approach to study the current problem of interest is described. As shown in Fig.\ref{Fig2}, this approach is composed of three major steps. The first two steps are used to consume high-fidelity simulation data and construct a model for an initial prediction of drag and lift forces. This is achieved through the integration of a dimension reduction method, ST-HOSVD (step 1), and a statistical modeling, GPR (step 2). The third step is to assimilate experimental measurement data to update the prediction by using ROPF (step 3). More details of each step will be elaborated in Subsections 4.1 to 4.3.

\subsection{Step 1: Sequentially-Truncated Higher-Order Singular Value Decomposition (ST-HOSVD) }

Here, ST-HOSVD, which is a multilinear extension of the matrix SVD to higher-order tensors, is employed to reduce the dimension of the offline simulation data $\mathbf{\chi}$ in different operation, temporal-like angle, and operation conditions. A truncation strategy is employed to reduce the computational cost without losing high accuracy. 

Let the tensor $\mathbf{\chi} \in \mathbb{R}^{I \times J \times T}$ represent an ensemble of offline simulated data sets. Here, $I$ is number of operation conditions, $J$ is number of system behavior (e.g. drag and lift forces) of prediction interests, $T$ is number of temporal-like angles. Here, a three-mode SVD is used to decompose the tensor data $\mathbf{\chi}$, as shown in:
\begin{equation}
\begin{array}{lcl}
\displaystyle\mathbf{\chi}&=&\displaystyle \xi x_{1} \mathbf{U} x_{2} \mathbf{V} x_{3} \mathbf{W}\\[6pt]
\end{array}\vspace*{-12pt}
\end{equation}
\vspace*{-3pt}

where $x_{1} \mathbf{U}$, $x_{2} \mathbf{V}$  and $x_{3} \mathbf{W}$  denote multiplication of the core tensor $\xi \in \mathbb{R}^{I \times J \times T}$  and the eigen design matrices $\mathbf{U} \in \mathbb{R}^{I \times I }$,  eigen operation matrices $\mathbf{V} \in \mathbb{R}^{J \times J}$ , and eigen temporal matrices $\mathbf{W} \in \mathbb{R}^{T \times T}$ , form the first, second and third indices of $\xi$ with the second indices of $\mathbf{U}$,  $\mathbf{V}$ and $\mathbf{W}$, respectively. Here, the core tensor $\xi$ governs the interaction between the features of robotic appendage design and operation along time in the eigen matrices. 

To reduce the dimension of the tensor data $\chi$, the multilinear eigenvectors and slices of the core tensors can be discarded based on the percentage of information desired to be kept. This is done by computing the multilinear eigen values and determining the rank of the tensor. Here, the ranks for the multi-dimensional tensor data can be different along each dimension. Let $r$ be the respective ranks of tensor $\chi$ selected for the temporal dimensions. Thus, the truncated eigen temporal matrices are $\widetilde{\mathbf{W}} \in \mathbb{R}^{T \times r }$. By a multilinear projection of the tensor data $\chi$ onto the truncated eigen temporal matrices, a set of low-dimensional parameter data $\mathcal{A}\in \mathbb{R}^{I \times J \times r}$ can be obtained by:
\begin{equation}
\begin{array}{lcl}
\displaystyle \mathcal{A}& = &\displaystyle \xi x_{3} \widetilde{\mathbf{W}}\\[6pt]
\end{array}\vspace*{-12pt}
\end{equation}
\vspace*{-3pt}
Each parameter matrix contains a set of coefficients for eigen design and operation temporal matrices, which represents the interaction between robotic appendage design and operation variability. 
\subsection{Step 2: Gaussian Process Regression (GPR)}
To model the variability of the temporal-like responses with respect to the appendage design and operation, a matrix of random variables, $\boldsymbol{\alpha} \in \mathbb{R}^{r_1}$, is defined to statistically model the projected parameters. Through an integration of these random variables with eigen operation and temporal-like matrices, the system’s temporal-like output can be approximated, as in the following equation:
\begin{equation}
\begin{array}{lcl}
\displaystyle f(\mathbf{x},c, \theta) & \cong &\displaystyle \sum^{r_1}_{j=1}\sum^{r_2}_{k=k_1}\alpha V_j^T \mathbf{\otimes} w_k^T \\[6pt]
\end{array}\vspace*{-12pt}
\end{equation}
\vspace*{-3pt}
Here, $\alpha_{j,k}$ is a component of the parameter matrix $\boldsymbol{\alpha}$ with respect to each of its indices. Each of these parameter components is independently modeled through the GPR $\boldsymbol{\alpha} =\mathbf{g}(\mathbf{x},c)$, which are trained by the estimated parameter data set obtained from Step 1. 

\subsection{Step 3: Reduced-Order Particle Filter (ROPF) }
In this paper, the ROPF method, proposed in the authors’ previous work \cite{Zhao_1}, is employed to update the system’s temporal estimates with a limited number of experimental measurement data. To do this, using ROPF, one takes advantage of the eigen temporal-like matrices. Instead of inferring the distribution of the entire system behavior, the authors estimate the posterior distribution of the projected parameter elements given the sensors measurement data, and thus this can significantly reduce the dimension and computational cost of the assimilation process.

\section{Results and Discussions}
\label{Results and Discussions}

\begin{table*}[htbp]
\centering
\caption{Design and operation scenarios for simulation data}
\label{tab:1} 

\begin{tabular}{lllllllllllllllllllllllllllll}
\hline\noalign{\smallskip}
Robotic appendage design&     Operation condition&   \\ 
\noalign{\smallskip}
\hline\noalign{\smallskip}
Flat leg: fixed leg length& Varying stride frequency &                    \\ 
\noalign{\smallskip}
\hline\noalign{\smallskip}
C-leg: fixed leg radius&   Varying stride frequency &        \\
\noalign{\smallskip}
\hline\noalign{\smallskip}
Reversed C-leg: fixed leg radius&              Varying stride frequency &        \\
\noalign{\smallskip}
\hline\noalign{\smallskip}
L-leg: fixed leg length&              Varying stride frequency &        \\
\noalign{\smallskip}
\hline\noalign{\smallskip}
Reversed L-leg: fixed leg length& Varying stride frequency &        \\
\noalign{\smallskip}
\hline\noalign{\smallskip}
L-leg: varying foot length& Fixed stride frequency  &        \\
\noalign{\smallskip}
\hline\noalign{\smallskip}
Reversed L-leg: varying foot length& Fixed stride frequency  &        \\
\noalign{\smallskip}
\hline
\end{tabular}
\end{table*}

\begin{figure*}[bthp]
	\includegraphics[width=1\textwidth]{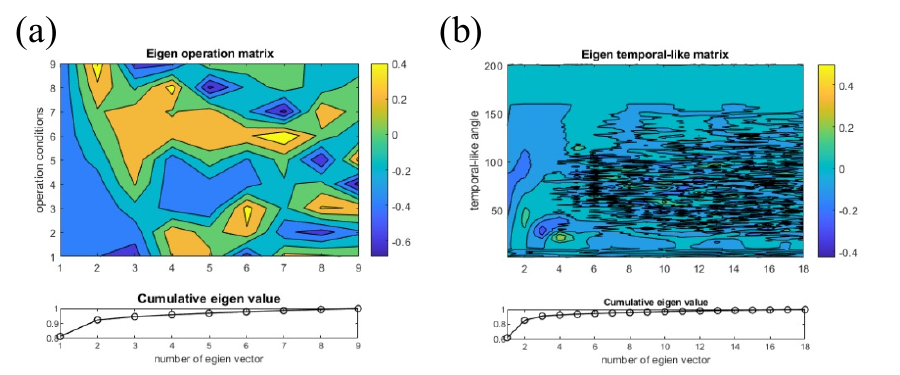}
	\caption{\label{Fig3} (a) Eigen operation matrices (b) Eigen temporal matrices.}
\end{figure*}
%

\begin{figure*}[th]
	\includegraphics[width=1\textwidth]{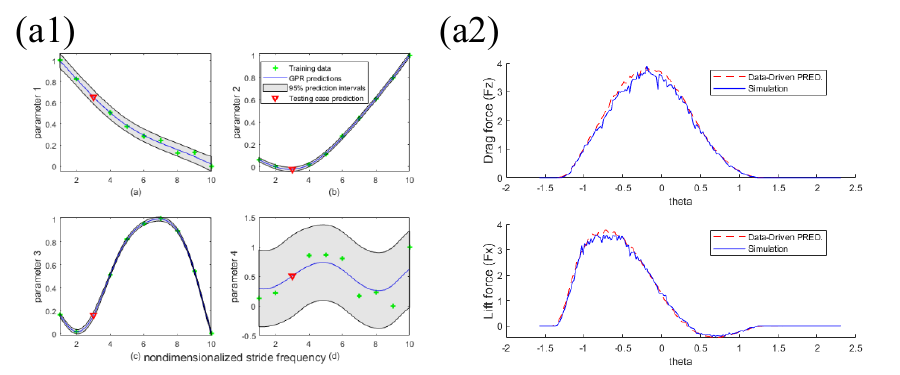}
	\caption{\label{Fig4}(a1) Gaussian process regression for low dimensional parameters for test case (a2) Comparison of resistive forces between simulation and prediction. }
\end{figure*}

\begin{figure*}[bthp]
	\centering  \includegraphics[width=1\textwidth]{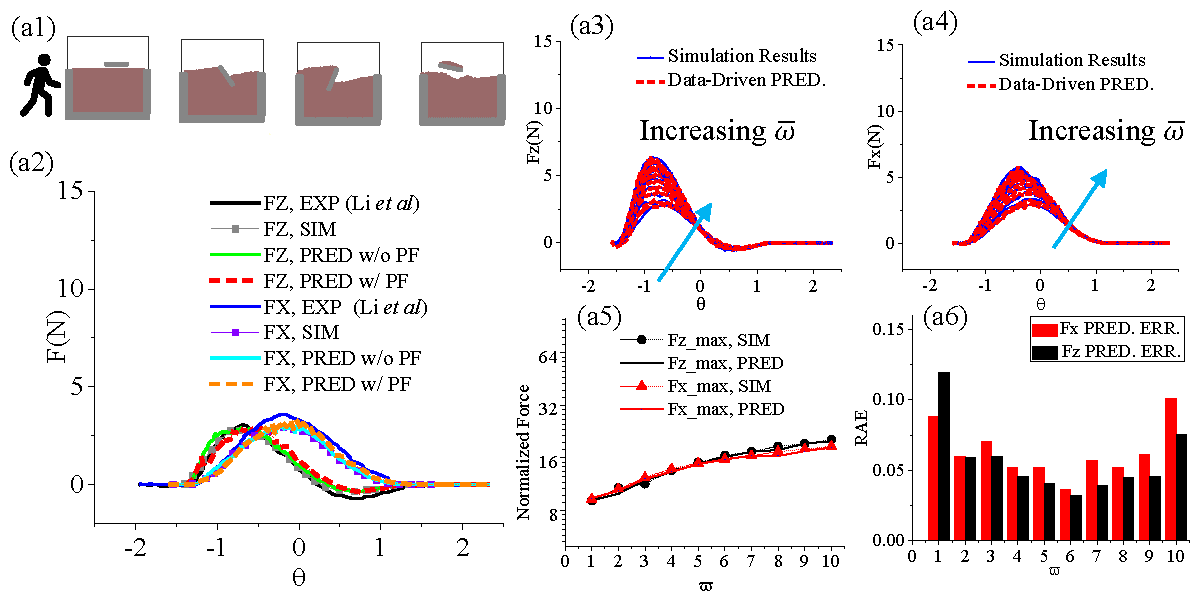}
	\centering \caption{\label{Fig5}Comparison of data driven prediction and physics-based simulation for flat leg, (a1) Real life leg morphology and profiles at different representative time, (a2) Comparison of net forces from experiments, simulations, data-driven prediction with particle filters, data-driven prediction without particle filters, (a3) Net lift Fz, (a4) Net thrust Fx, (a5) Sensitive dependence of normalized net forces on $ \varpi $, (a6) Relative absolute error for net forces prediction.}
	
\end{figure*}

The effectiveness of the data-driven modeling framework will be demonstrated by predicting the behavior of rigid robot leg interaction with the granular material. The motion of the rigid body is prescribed to be consistent with the experiment, which is a prototype of the legged robot. In this case study, two sources of data are available from physics-based simulation and physical experiments.

Physics-based simulation  will be used to collect a set of data for different robotic appendage designs and operation conditions. Here, seven cases of appendage designs and operations are studied as illustrated in Table \ref{tab:1}, where the physical parameters for each robot leg can be found in \cite{Li_2}. Those robot legs are interacting on granular media, which are closely packed 3 mm (diameter) glass spheres with a particle material density of $2.6g/cm^3$ and angle of repose $23^o$. During the robot leg rotation simulation, the leg is first paused for 2 seconds, and then rotated clockwise at a hip height 2cm within leg angle $-3\pi/4 \le  \theta \le 3\pi/4$. For these combinations of robot leg design and operation, simulations are conducted to simulate the drag and lift forces. Physical experiments \cite{Li_2} are only available for three subsets of design and operation combinations, which are the flag leg, C-leg, and reverse C-leg with stride frequency as $0.2 rad/s$.

In this case study, first, a step-by step procedure of the proposed approach will be used to predict behaviors for the flag robot leg design on several operation conditions. Then, results for all robot leg designs and discussions will be presented.

\subsection{ Application of Proposed Approach on Flat Robot Leg Design}


The flat robot leg design and its real-life leg morphology can be visualized in Fig.\ref{Fig5} (a1), where the operating profiles for such flat leg at different representative times can be also gleaned. As shown in Table 1, 10 operation conditions (stride frequency) are considered for generating data for the flag robot leg. For testing the algorithm, a ten-fold cross-validation algorithm is used. That is, each of the ten operating conditions is used as a testing data, while the other nine scenarios are used as training data sets. The example of the step-by-step procedure will be provided on one of the operating conditions at stride frequency 0.6 rad/s.

First, ST-HOSVD is used to decomposed the training data sets into eigen operation and temporal-like matrices, as plotted in Fig.\ref{Fig3} (a) and Fig.\ref{Fig3} (b), respectively. It is composed of a set of dominant eigen vectors which capture variation of system behavior vary along different operating condition and temporal-like $\theta$. Meanwhile, for each eigen matrices, the cumulative eigen value versus number of eigen vector can also be visualized in Fig.\ref{Fig3}. In this case study, the threshold for kept information is set as 95$\%$, then ranks for operation and temporal-like dimensions have been determined as 4 and 7, respectively. Next, by projecting the simulation data onto the eigen matrices, a low-dimensional parameter $\mathbb{R}^{9x4}$ data set can be obtained. 

Second, four GPR models have been constructed to model the variation of four low-dimensional parameters under different operation conditions, as shown in Fig.\ref{Fig4}. Each projected parameters at different operating conditions (stride frequencies) are plotted as green cross in Fig.\ref{Fig4}. The GPR model predictions and its 95$\%$ prediction confidence intervals are visualized as blue lines and grey areas in Fig.\ref{Fig4}. Then, such GPR models are used to predict four parameters at the test case, where the stride frequency is equal to 0.6 rad/s. The predicted values are indicated by red triangles in Fig.\ref{Fig4}. With the integration of predicted parameters and eigen operation and temporal-like matrices, the estimation of drag and lift forces can be obtained as shown in Fig.\ref{Fig4}, which has very comparable accuracy to results from high-fidelity simulation.

A similar application of the data-driven model procedure is applied to predict the drag and lift forces for each of the 10 operating conditions. In Fig.\ref{Fig5}, overall, the simulation of robot leg rotation is shown in Fig.\ref{Fig5} (a1). The predicted drag and lift forces with nondimensionalized stride frequencies ranging from 1 to 10 are illustrated in in Fig.\ref{Fig5} (a3) and (a4), respectively. It shows that the data-driven prediction approach can generate predictions that have comparable accuracy as simulations across different stride frequencies using only simulation data as training data.

On the other hand, experimental data are available for the case where the non-dimensionalized stride frequency is 1. ROPF is used to further update the prediction by assimilating the sparse experiment data. Comparison of drag and lift forces have been provided in Fig.\ref{Fig5} (a2), which includes experiment data, simulation data, data-driven prediction with and without PF approach. As can be gleaned from Fig.\ref{Fig5} (a2), the predictions match well with both the simulation and experiment, and even outperform the simulation to be closer to the experiment. Additionally, the average absolute errors between simulation and prediction for all 10 operation conditions are calculated as shown in Fig.\ref{Fig5} (a2).

\subsection{Prediction for Robot Appendage Interaction with Granular Terrain across Leg Morphology }

\begin{figure*}[htbp]
	\includegraphics[width=1\textwidth]{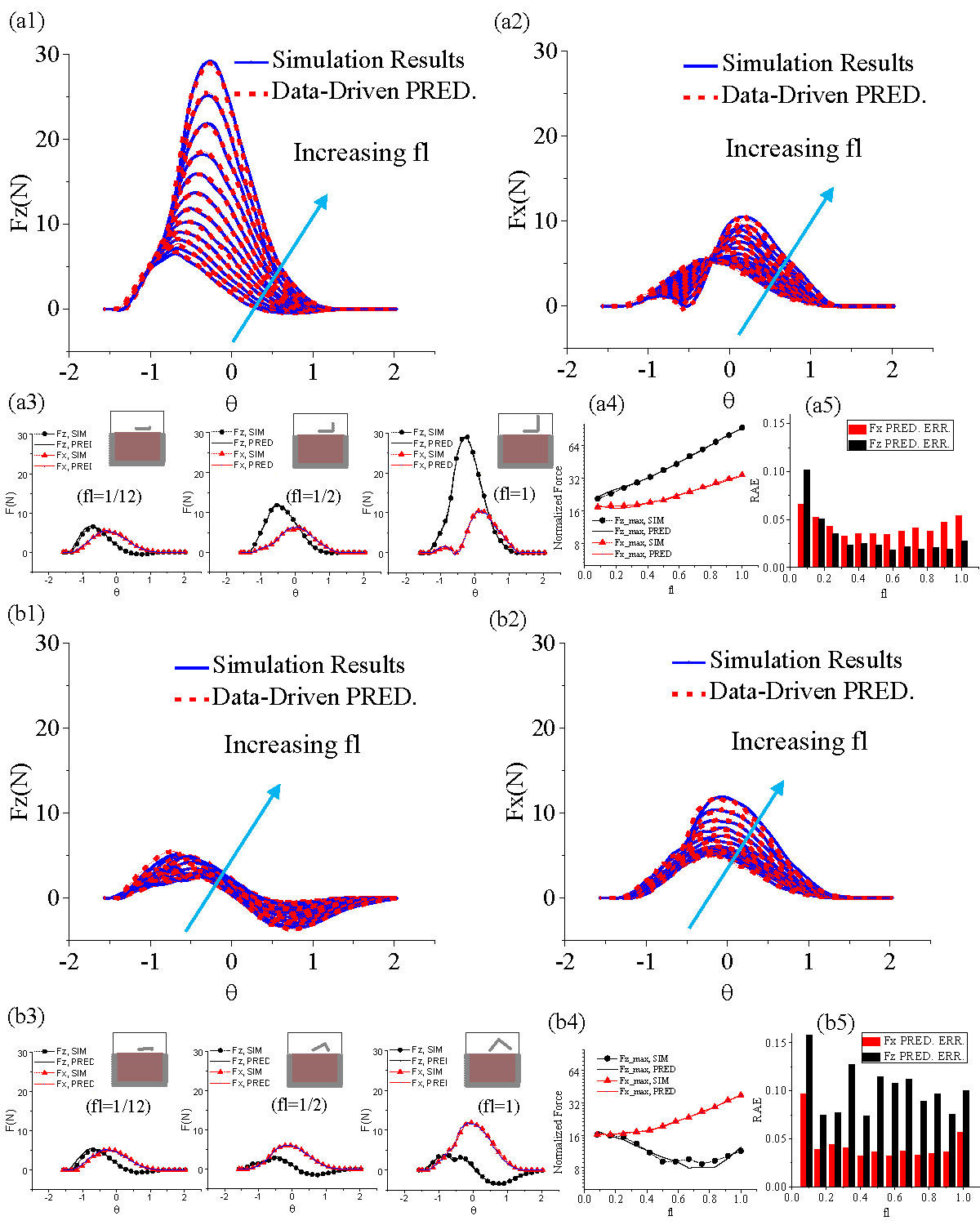}
	\caption{\label{Fig6}Comparison of data driven prediction and physics-based simulation for L-leg and Reversed L-leg across different foot lengths. a) L-leg: (a1) Net lift Fz, (a2) Net thrust Fx, (a3) Net forces for fl=1/12, fl=1/2 and fl=1, (a4) Sensitive dependence of normalized net forces on fl (a5) Relative absolute error for net forces prediction. b) Reversed L-leg: (b1) Net lift Fz, (b2) Net thrust Fx, (b3) Net forces for fl=1/12, fl=1/2 and fl=1, (b4) Sensitive dependence of normalized net forces on fl (b5) Relative absolute error for net forces prediction.   }	
\end{figure*}

\begin{figure*}[htbp]
	\includegraphics[width=1\textwidth]{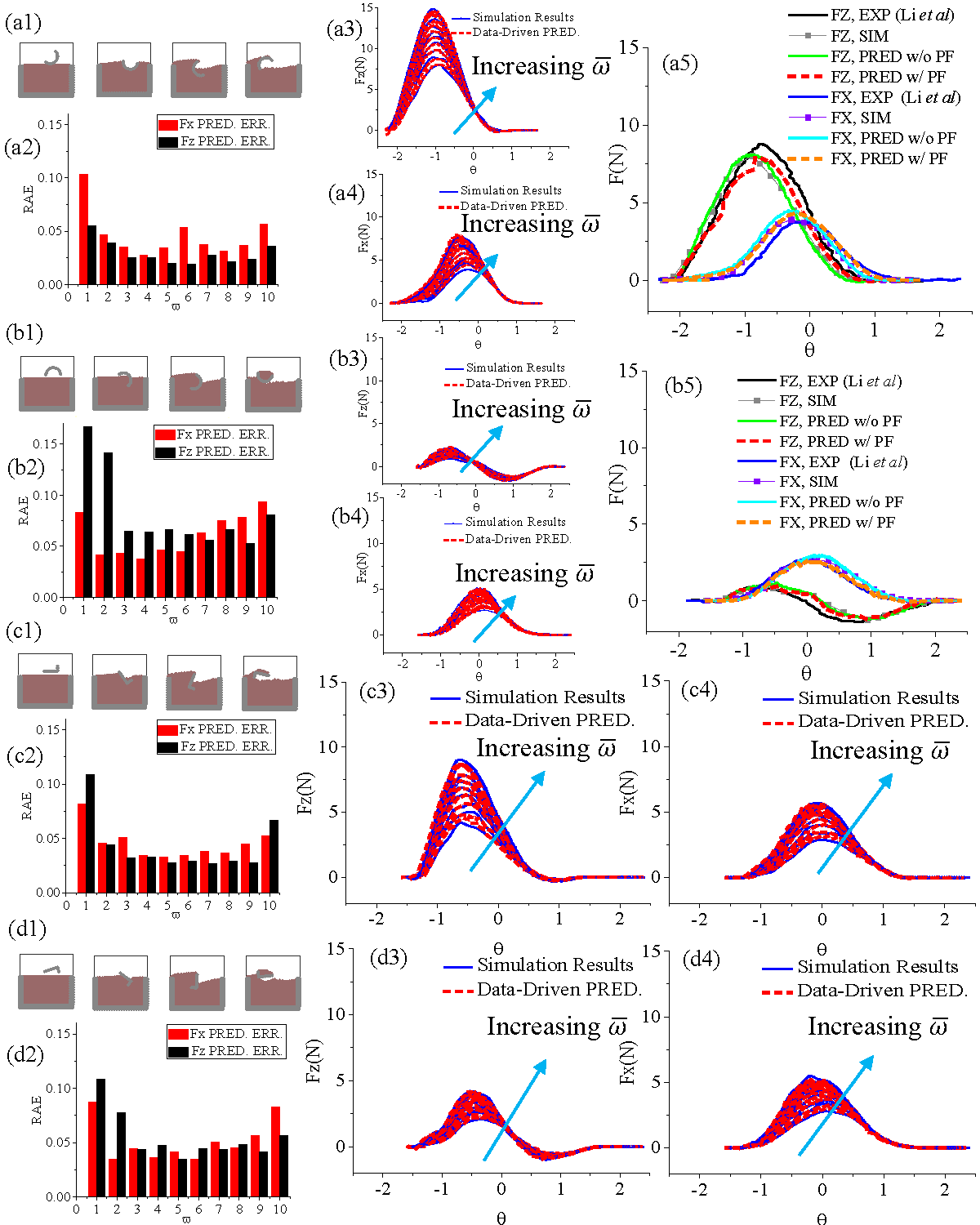}
	\caption{Comparison of data driven prediction and physics-based simulation for C-leg, Reversed C-leg, L-leg and Reversed L-leg across different rotation speeds. a) C- leg: (a1) Profiles at different representative time, (a2) Relative absolute error for net forces prediction, (a3) Net lift Fz prediction, (a4) Net thrust Fx prediction, (a5) Comparison of net forces from experiments, simulations, data-driven prediction with particle filters, data-driven prediction without particle filters. b) Reversed C- leg. c) L-leg, fl=1/3. d) Reversed L-leg, fl=1/3.}
	\label{Fig7}
\end{figure*}
%

\begin{figure*}[htbp]
	\includegraphics[width=1\textwidth]{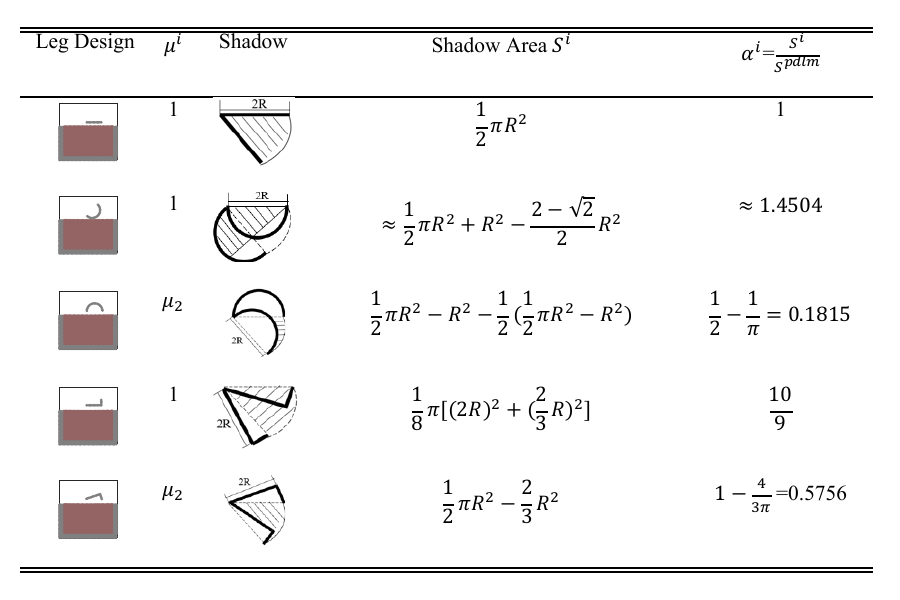}
	\caption{Scale Factor Across Leg Morphologies.}
	\label{Fig8} 
\end{figure*}
%
\begin{figure*}[htbp]
	\includegraphics[width=1\textwidth]{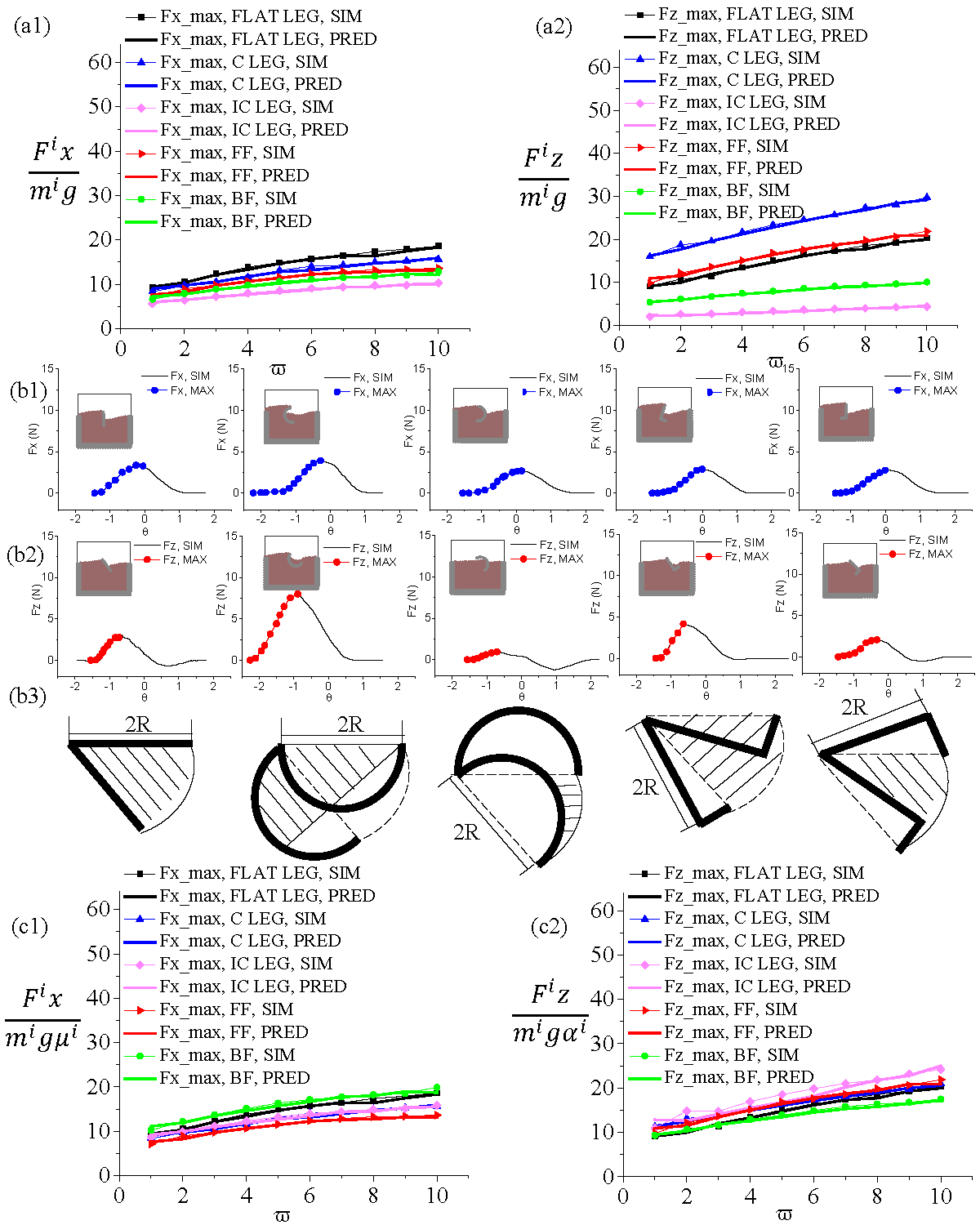}
	\caption{Comparison of data driven prediction and physics-based simulation for sensitive dependence of normalized net forces on rotation speeds $ \varpi $ for different leg appendages. (a1) Sensitive dependence of maximum normalized drag forces on $ \varpi $, (a2) Sensitive dependence of maximum normalized lift forces on $ \varpi $, (b1) Profiles before and when drag forces reach maximum, (b2) Profiles before and when lift forces reach maximum, (b3) Approximate granular material area (volume) that legs lift (c1) Sensitive dependence of renormalized drag forces on $ \varpi $, (c2) Sensitive dependence of renormalized lift forces on $ \varpi $}
	\label{Fig9} 
\end{figure*}
Following the same application, the proposed data-driven approach has been evaluated on all other six robotic appendage designs from Table \ref{tab:1}. Comparison between data-driven prediction and simulation is conducted for L-leg (Fig.\ref{Fig6} (a1)-(a5)) and reverse L-leg (Fig.\ref{Fig6} (b1)-(b6)) across different foot lengths. As shown in these figures, the proposed approach has demonstrated the capability to capture the trend of resistive forces with the increase of foot lengths. From Fig.\ref{Fig6} (a1) to (a4) and Fig.\ref{Fig6} (b1) to (b4), an observation is that the lift force of L-leg increases with longer foot, however, that of Reversed L-leg first decreases then increases with longer foot. As the trend of lift force for Reversed L-leg is more complex than L-leg, from Fig.\ref{Fig6} (a5) and (b5), the error of the lift force (Fz) prediction of Reversed L-leg is larger than that of L-leg, indicating more data is needed for a better prediction of lift force for Reversed L-leg. 

Additionally, further comparison of data driven prediction and physics-based simulation are provided for C-leg (Fig.\ref{Fig7} (a1)-(a5)), Reverse C-leg (Fig.\ref{Fig7} (b1)-(b5)), L-leg (Fig.\ref{Fig7} (c1)-(c5)), and Reversed L-leg (Fig.\ref{Fig7} (d1)-(d5)) across different rotation speeds. As shown in these figures, the proposed approach has demonstrated the capability to capture the dynamic interaction for different design of robot legs under different operating conditions. In all the cases, the errors in the prediction of edge cases are larger than those in between. This is because for the cases in between, data from both sides are used in the prediction; for edge cases, only data from one side are leveraged. 

The prediction mentioned above has a case-by-case basis, which means that for a fixed robot leg design, the prediction can be made under different operating conditions. However, predictions cannot be made across various designs. As a trail to develop a unified prediction framework, we recovered a scaling relationship for maximum resistive forces with different rotation speeds across different designs. For drag forces, the scale factor $\mu^i$  is equal to 1 or dynamic friction coefficient, which is the limiting value of friction coefficient ($\mu_2=tan33^o$ ) at high inertial number in $\mu(I)$ rheology. For lift forces, the scale factor $\alpha^i$ depends on the volume (area) of granular material the leg lift since the lift force is buoyancy like force. The scale factors for different leg morphologies can be calculated from Fig.\ref{Fig8}.  

The sensitive dependence of normalized net forces on rotation speeds is shown in Fig.\ref{Fig9}. In (a1), without introducing the scale factors, the relationship between net forces and rotation speeds across different designs cannot be unified. Based on our observation from the simulations shown in (b1), the drag forces reach maximum when reference lines (diameters for C and IC legs, hypotenuses for L and IL legs) are close to vertical. For flat leg, C leg and L leg, the drag forces are dominated by pressure, while for IC leg and IL leg, the drag forces are dominated by shear force. Therefore, the scale factor $\mu^i$  is set to 1 for flat leg, C leg and L leg, and dynamic friction coefficient for IC leg and IL leg. From what we observed in simulations shown in (b2), the lift forces reach maximum when reference lines are close to 45 degrees to the horizontal. As the lift forces are buoyancy like forces, the volume (area) of granular material the leg lift is defined as $S^i$, shown as the shadow area in (b3), where i stands for different leg morphologies. The scale factor is approximated as the ratio between Si and $S^{pdlm}$, where $S^{pdlm}$ is the volume (area) of granular material the flat leg lifts. After introducing these scale factors, the sensitive dependence of normalized net forces on rotation speeds can be unified shown in (c1) and (c2).

\section{Concluding Remarks}
\label{Concluding Remarks}

The authors provided an alternative approach for effective prediction of the interaction of the robotic appendage with granular materials. With the unification of dimension reduction, surrogate modeling, and data assimilation techniques, the data-driven modeling approach can provide a quite reliable prediction by integrating data from high-fidelity simulations and sparse physical experiment. Significant reduction in computational time is achieved with the data-driven modeling approach compared with physics-based simulations. In addition, the data-driven approach has general predictability beyond a case-by-case basis and the potential of outperforming only simulations in the long-term predictions. 
Although the method showcases promising results for the current cases, it is not without its drawbacks. The accuracy of the current method drops when the trend of the data becomes complex. The prediction is less ideal mainly due to simulation and experiment data are only available in a very limited number of design scenarios. For further improve prediction performance across different robotic left designs, more data should be collected through simulation and experiment for various leg designs.



\begin{acknowledgements}
 The authors would like to thank the reviewers for the constructive suggestions provided during the review process. The work presented in this paper was supported in part by the U.S. National Science Foundation Grant No.1507612 and U.S. Air Force Office of Scientific Research through Grant No. FA9550150134. The authors gratefully acknowledge the computational resources provided by the Maryland Advanced Research Computing Center (MARCC). This study was conducted while the authors Dr. Guanjin Wang, as a postdoc, and Dr. Xiangxue Zhao, as a Ph.D. student, were at the University of Maryland, College Park. 
\end{acknowledgements}
\boldsymbol{$Declarations$}

\textbf{Conflict of interest} 
The authors have no competing interests to declare that are relevant to the content of this article.

\bibliographystyle{}

\end{document}